\def\year{2017}\relax
\documentclass[letterpaper]{article}
\usepackage{aaai18}
\usepackage{times}
\usepackage{helvet}
\usepackage{courier}
\usepackage{url}
\usepackage{graphicx}
\usepackage{amsmath,amssymb,amsfonts,amsthm}

\theoremstyle{plain}
\newtheorem{thm}{Theorem}
\newtheorem{lem}{Lemma}
\newtheorem{prop}{Proposition}

\RequirePackage{algorithm}
\RequirePackage[noend]{algpseudocode}

\renewcommand\[{\begin{equation}}
\renewcommand\]{\end{equation}}

\let\emptyset\varnothing

\newcommand{\bbR}{\mathbb{R}}

\newcommand{\calvar}[1]{\ensuremath{\mathcal{#1}}}

\newcommand{\calG}{\calvar{G}}

\newcommand{\calI}{\calvar{I}}
\newcommand{\calJ}{\calvar{J}}

\newcommand{\calO}{\calvar{O}}
\newcommand{\calP}{\calvar{P}}

\newcommand{\calX}{\calvar{X}}

\newcommand{\vecvar}[1]{\ensuremath{\boldsymbol{#1}}}
\newcommand{\va}{\vecvar{a}}
\newcommand{\vb}{\vecvar{b}}

\newcommand{\vw}{\vecvar{w}}
\newcommand{\vx}{\vecvar{x}}

\newcommand{\vphi}{\vecvar{\phi}}

\DeclareMathOperator*{\argmax}{argmax}
\DeclareMathOperator*{\argmin}{argmin}

\newcommand{\norm}[1]{\lVert#1\rVert}
\newcommand{\inner}[2]{\langle #1, #2 \rangle}

\usepackage{xcolor}

\newcommand{\pcl}{\textsc{pcl}}

\newcommand{\REG}[1]{\ensuremath{\text{REG$\left(#1\right)$}}}
\newcommand{\CREG}[2]{\ensuremath{\text{CREG}_{#1}\left(#2\right)}}
\newcommand{\ACREG}[1]{\ensuremath{\text{CREG}_{#1}}}

\title{Decomposition Strategies for Constructive Preference Elicitation \\ (and Supplementary Material)}
\author{Paolo Dragone\thanks{PD is a fellow of TIM-SKIL Trento and is supported
by a TIM scholarship. This work has received funding from the European Research
Council (ERC) under the European Union’s Horizon 2020 research and innovation
programme (grant agreement No.  [694980] SYNTH: Synthesising Inductive Data
Models).}\\
University of Trento, Italy\\
TIM-SKIL, Trento, Italy\\
\texttt{paolo.dragone@unitn.it}
\And
Stefano Teso
\\
KU Leuven, Belgium\\
\texttt{stefano.teso@cs.kuleuven.be}
\AND
Mohit Kumar$\,^\dagger$\\
KU Leuven, Belgium\\
\texttt{mohit.kumar@cs.kuleuven.be}
\And
Andrea Passerini\\
University of Trento, Italy\\
\texttt{andrea.passerini@unitn.it}
}

\begin{document}
\maketitle

\begin{abstract}
  We tackle the problem of constructive preference elicitation, that
  is the problem of learning user preferences over very large decision
  problems, involving a combinatorial space of possible outcomes. In
  this setting, the suggested configuration is synthesized on-the-fly
  by solving a constrained optimization problem, while the preferences
  are learned iteratively by interacting with the user. Previous work
  has shown that Coactive Learning is a suitable method for learning
  user preferences in constructive scenarios. In Coactive Learning the
  user provides feedback to the algorithm in the form of an
  improvement to a suggested configuration. When the problem involves
  many decision variables and constraints, this type of interaction
  poses a significant cognitive burden on the user. We propose a
  decomposition technique for large preference-based decision problems
  relying exclusively on inference and feedback over partial
  configurations.  This has the clear advantage of drastically
  reducing the user cognitive load. Additionally, part-wise inference
  can be (up to exponentially) less computationally demanding than
  inference over full configurations.  We discuss the theoretical
  implications of working with parts and present promising empirical
  results on one synthetic and two realistic constructive problems.
\end{abstract}

\section{Introduction}

In constructive preference elicitation (CPE) the recommender aims at suggesting
a \emph{custom} or \emph{novel} product to a
customer~\cite{teso2016constructive}. The product is assembled on-the-fly from
components or synthesized anew by solving a combinatorial optimization problem.
The suggested products should of course satisfy the customer's preferences,
which however are unobserved and must be learned
\emph{interactively}~\cite{pigozzi16preferences}. Learning proceeds
iteratively: the learner presents one or more candidate recommendations to the
customer, and employs the obtained feedback to estimate the customer's
preferences.  Applications include recommending custom PCs or
cars, suggesting touristic travel plans, designing room and building layouts,
and producing recipe modifications, among others.

A major weakness of existing CPE
methods~\cite{teso2016constructive,teso2017coactive} is that they require the
user to provide feedback on \emph{complete} configurations. In real-world
constructive problems such as trip planning and layout design, configurations
can be large and complex. When asked to evaluate or manipulate a complex
product, the user may become overwhelmed and confused, compromising the
reliability of the obtained feedback~\cite{mayer2003nine}. Human decision
makers can revert to a potentially uninformative prior when problem solving
exceeds their available resources. This effect was observed in users tasked
with solving simple SAT instances (three variables and eight
clauses)~\cite{ortega2016human}.  In comparison, even simple constructive
problems can involve tens of categorical variables and features, in addition to
hard feasibility constraints.  On the computational side, working
with complete configurations poses scalability problems as well. The reason is
that, in order to select recommendations and queries, constructive recommenders
employ constraint optimization techniques. Clearly, optimization of complete
configurations in large constructive problems can become computationally
impractical as the problem size increases.

Here we propose to exploit factorized utility
functions~\cite{braziunas2009elicitation}, which occur very naturally in
constructive problems, to work with \emph{partial} configurations. In
particular, we show how to generalize Coactive Learning
(CL)~\cite{shivaswamy2015coactive} to part-wise inference and learning. CL is a
simple, theoretically grounded algorithm for online learning and preference
elicitation. It employs a very natural interaction protocol: at each iteration
the user is presented with a single, appropriately chosen candidate
configuration and asked to improve it (even slightly).
In~\cite{teso2017coactive}, it was shown that CL can be lifted to constructive
problems by combining it with a constraint optimization solver to efficiently
select the candidate recommendation. Notably, the theoretical guarantees of CL
remain intact in the constructive case.

Our part-wise generalization of CL, dubbed \pcl, solves the two aforementioned
problems in one go: (i) by presenting the user with partial configurations, it
is possible to (substantially) lessen her cognitive load, improving the
reliability of the feedback and enabling learning in larger constructive tasks;
(ii) in combinatorial constructive problems, performing inference on partial
configurations can be \emph{exponentially} faster than on complete ones.
Further, despite being limited to working with partial configurations, \pcl\
can be shown to still provide \emph{local optimality} guarantees in theory, and
to perform well in practice.

This paper is structured as follows. In the next section we overview
the relevant literature. We present \pcl\ in the Method section,
followed by a theoretical analysis. The performance of \pcl\ are then
illustrated empirically on one synthetic and two realistic constructive
problems. We close the paper with some concluding remarks.


\section{Related Work}

Generalized additive independent (GAI) utilities have been thoroughly explored
in the decision making literature~\cite{fishburn1967interdependence}. They
define a clear factorization mechanism, and offer a good trade off between
expressiveness and ease of
elicitation~\cite{chajewska2000making,gonzales2004gai,braziunas2009elicitation}.
Most of the early work on GAI utility elicitation is based on graphical models,
e.g. UCP and GAI networks~\cite{gonzales2004gai,boutilier2001ucp}. These
approaches aim at eliciting the full utility function and rely on the
comparison of full outcomes. Both of these are infeasible when the utility
involves many attributes and features, as in realistic constructive problems.

Like our method, more recent
alternatives~\cite{braziunas2005local,braziunas2007minimax} handle both partial
elicitation, i.e. the ability of providing recommendations without full utility
information, and local queries, i.e. elicitation of preference information by
comparing only (mostly) partial outcomes.
There exist both Bayesian~\cite{braziunas2005local} and
regret-based~\cite{braziunas2007minimax,boutilier2006constraint} approaches,
which have different shortcomings. Bayesian methods do not scale
to even small size constructive problems~\cite{teso2016constructive},
such as those occurring when reasoning over individual parts in
constructive settings. On the
other hand, regret-based methods require the user feedback to be strictly
self-consistent, an unrealistic assumption when interacting with non-experts.
Our approach instead is specifically designed to scale to larger constructive
problems and, being derived from Coactive Learning, natively handles
inconsistent feedback.
Crucially, unlike \pcl, these local elicitation methods also require to perform
a number of queries over \emph{complete} configurations to calibrate the
learned utility function. In larger constructive domains this is both
impractical (on the user side) and computationally infeasible (on the learner
side).

Our work is based on Coactive Learning (CL)~\cite{shivaswamy2015coactive}, a
framework for learning utility functions over structured domains, which has been
successfully applied to CPE~\cite{teso2017coactive,dragone2016layout}. When applied to constructive
problems, a crucial limitation of CL is that the learner and the user interact
by exchanging \emph{complete} configurations. Alas, inferring a full
configuration in a constructive problem can be computationally demanding,
thus preventing the elicitation procedure from being real-time. This can be
partially addressed by performing approximate inference, as
in~\cite{raman2012online}, at the cost of weaker learning guarantees. A
different approach has been taken in~\cite{goetschalckx2014coactive}, where the
exchanged (complete) configurations are only required to be locally optimal, for
improved efficiency.  Like \pcl, this method guarantees the local
optimality of the recommended configuration.  All of the previous approaches,
however, require the user to improve a potentially large \emph{complete}
configuration. This is a cognitively demanding task which can become prohibitive
in large constructive problems, even for domain experts, thus
hindering feedback quality and effective elicitation. By dealing with parts
only, \pcl\ avoids this issue entirely.


\section{Method}

\paragraph{Notation.} We use rather standard notation: scalars $x$ are written
in italics and column vectors $\vx$ in bold. The inner product of two vectors
is indicated as $\langle \va, \vb \rangle = \sum_i a_i b_i$, the Euclidean norm
as $\|\va\| = \sqrt{\langle \va, \va \rangle}$ and the max-norm as
$\|\va\|_\infty = \max_i a_i$. We abbreviate $\{1, \ldots, n\}$ to $[n]$, and
indicate the complement of $I \subseteq [n]$ as $-I := [n]\,\backslash\,I$.

\paragraph{Setting.} Let $\calX$ be the set of candidate structured products.
Contrarily to what happens in standard preference elicitation, in the
constructive case $\calX$ is defined by a set of hard constraints rather than
explicitly enumerated\footnote{In this paper, ``hard'' constraints refer to the
constraints delimiting the space of feasible configurations, as opposed to
``soft'' constraints, which determine preferences over feasible
configurations~\cite{meseguer2006soft}.}.  Products are represented by a
function $\vphi(\cdot)$
that maps them to an $m$-dimensional feature space. While the feature map can
be arbitrary, in practice we will stick to features that can be encoded as
constraints in a mixed-integer linear programming problem, for efficiency; see
the Empirical Analysis section for details.  We only assume that the features
are bounded, i.e.  $\forall x \in \calX$ it holds that $\|\vphi(x)\|_{\infty}
\le D$ for some fixed $D$.

As is common in multi-attribute decision theory~\cite{keeney1976}, we assume
the desirability of a product $x$ to be given by a utility function $u^*: \calX
\to \bbR$ that is linear \emph{in the features}, i.e.,
$u^*(x) := \langle \vw^*, \vphi(x) \rangle$.
Here the weights $\vw^* \in \bbR^m$ encode the true user preferences, and may
be positive, negative, or zero (which means that the corresponding feature is
irrelevant for the user). Utilities of this kind can naturally express rich
constructive problems~\cite{teso2016constructive,teso2017coactive}.

\paragraph{Parts.} Here we formalize what parts and partial configurations are,
and how they can be manipulated. We assume to be given a set of $n$
\textit{basic parts} $p \in \calP$. A \textit{part} is any subset
$P \subseteq \calP$ of the set of basic parts.  Given a part $P$ and an object
$x$, $x_P\in\calX_P$ indicates the \textit{partial configuration} corresponding
to $P$. We require that the union of the basic parts reconstructs the whole
object, i.e. $x_{\calP} = x$ for all $x\in\calX$. The proper semantics of the
decomposition into basic parts is task-specific. For instance, in a scheduling
problem a month may be decomposed into days, while in interior design a house
may be decomposed into rooms. Analogously, the non-basic parts could then be
weeks or floors, respectively. In general, any combination of basic parts is
allowed. We capture the notion of combination of
partial configurations with the \textit{part combination} operator $\circ :
\calX_P \times \calX_Q \to \calX_{P \cup Q}$, so that $x_P \circ x_Q = x_{P \cup
Q}$. We denote the complement of part $P$ as $\overline{P} = \calP \setminus P$,
which satisfies $x = x_P \circ x_{\overline{P}}$ for all $x\in\calX$.

Each basic part $p\in\calP$ is associated to a \textit{feature subset} $I_p\subseteq[m]$, which
contains all those features that depend on $p$ (and only those). In general,
the sets $I_p$ may overlap, but we do require each basic part $p$ to be
associated to some features that do not depend on any other basic part $q$,
i.e.  that $I_p \setminus \left( \bigcup_{q \ne p} I_q \right) \neq \emptyset$
for all $p \in \calP$.  The features associated to a part $P\subseteq\calP$ are
defined as $\bigcup_{p \in P} I_p$. Since the union of the basic parts makes up
the full object, we also have that $\bigcup_{p \in \calP} I_p = [m]$.

\paragraph{GAI utility decomposition.} In the previous section we introduced a
decomposition of configurations into parts. In order to elicit the user
preferences via part-wise interaction, which is our ultimate goal, we need to
decompose the utility function as well. Given a part $P$ and its feature subset
$I_P$, let its \textit{partial utility} be:
\[ \textstyle u[I_P](x) := \inner{\vw_{I_P}}{\vphi_{I_P}(x)} = \sum_{i \in I_P} w_i \phi_i(x) \label{eq:partialutility} \]
If the basic parts have no shared features, the utility function is
\textit{additive}: it is easy to verify that
$u(x) = \sum_{p\in\calP} u[I_p](x)$. In this case, each part can be
managed independently of the others, and the overall configuration
maximizing the utility can be obtained by separately maximizing each
partial utility and combining the resulting part-wise configurations.

However, in many applications of interest the feature subsets
\textit{do} overlap. In a travel plan, for instance, one can be
interested in alternating cultural and leisure activities in
consecutive days, in order to make the experience more diverse and
enjoyable. In this case, the above decomposition does not apply
anymore as the basic parts may depend on each other through the shared
features. Nonetheless, it can be shown that our utility function is
\textit{generalized additive independent} (GAI) over the feature
subsets $I_p$ of the basic parts. Formally, a utility $u(x)$ is GAI if
and only if, given $n$ feature subsets
$I_1, \dots, I_n \subseteq [m]$, it can be decomposed into $n$
\textit{independent} subutilities~\cite{braziunas2005local}:
\[ u(x) = u_{I_1}(x) + \dots + u_{I_n}(x) \label{eq:gai} \]
where each subutility $u_{I_k}$ can only depend on the features in
$I_k$ (but does not need to depend on {\em all} of them). This
decomposition enables applying ideas from the GAI literature to
produce a well-defined part-wise elicitation protocol. Intuitively, we
will assign features to subutilities so that whenever a feature is
shared by multiple parts, only the subutility corresponding to one of
them will depend on that feature.

We will now construct a suitable decomposition of $u(x)$ into $n$
independent subutilities. Fix some order of the basic parts
$p_1, \dots, p_n$, and let:
\[ \textstyle J_k := I_k \setminus (\bigcup_{j = k+1}^n I_j) \label{eq:featdecomp} \]
for all $k\in[n]$. We define the subutilities as
$u_{I_k}(x) := u[J_k](x)$ for all $k \in [n]$. By summing up the
subutilities for all parts, we obtain a utility where each feature is
computed exactly once, thus recovering the full utility $u(x)$:
\begin{align*}
u(x) & = \textstyle \sum_{k=1}^n u_{I_k}(x)  = \textstyle \sum_{k=1}^n u[I_k \setminus \bigcup_{j = k+1}^n I_j](x) \\
& = \textstyle \sum_{i \in I_\calP} w_i \phi_i(x)
\end{align*}

\begin{algorithm}[t]
    \caption{\label{alg:ordselect} An example of ordering selection procedure using a GAI network~\cite{gonzales2004gai}.}
    \begin{algorithmic}[1]
        \Procedure{\textsc{SelectOrdering}}{$\calP$}
            \State Build a GAI network $\calG$ from $I_p$, $p \in \calP$
            \State Produce sequence $p_1, \dots, p_n$ by sorting the nodes \\
                   \qquad in $\calG$ in ascending order of node degree \\ \quad\ \
            \Return $p_1, \dots, p_n$
        \EndProcedure
    \end{algorithmic}
\end{algorithm}

The GAI decomposition allows to elicit each subutility $u_{I_k}$
separately.
By doing so, however, we end up ignoring some of the dependencies
between parts, namely the features in $I_k \setminus J_k$. This is the
price to pay in order to achieve decomposition and partwise
elicitation, and it may lead to suboptimal solutions if too many
dependencies are ignored. It is therefore important to minimize the
broken dependencies by an appropriate ordering of the parts. Going
back to the travel planning with diversifying features example, consider
a multi-day trip. Here the parts may refer to individual days, and
$I_k$ includes \emph{all} features of day $k$, including the features
relating it to the other days, e.g. the alternation of cultural and
leisure activities. Note that the $I_k$'s overlap.
On the other hand, the $J_k$'s are subset of features chosen so that
every feature only appears once. A diversifying feature relating
days 3 and 4 of the trip is either assigned to $J_3$ or $J_4$, but
not both.

One way to control the ignored dependencies is by leveraging GAI
networks~\cite{gonzales2004gai}.  A GAI network is a graph whose nodes
represent the subsets $I_k$ and whose edges connect nodes sharing at
least one feature. Algorithm~\ref{alg:ordselect} presents a simple and
effective solution to provide an ordering. It builds a GAI network
from $\calP$ and sorts the basic parts in ascending order of node
degree (number of incoming and outgoing edges). By ordering last the
subsets having intersections with many other parts, this ordering
attempts to minimize the lost dependencies in the above decomposition
(Eq.~\ref{eq:featdecomp}). This is one possible way to order the
parts, which we use as an example; more informed or task-specific
approaches could be devised.

\begin{algorithm*}[t]
    \caption{\label{alg:pcl} The \pcl\ algorithm.}
    \algnewcommand{\IIf}[1]{\State\algorithmicif\ #1\ \algorithmicthen}
    \algnewcommand{\IElse}{\unskip\ \algorithmicelse\ }
    \algnewcommand{\EndIIf}{\unskip\ ; }
    \begin{algorithmic}[1]
        \Procedure{\pcl}{$\calP$, $T$}
            \State $p_1, \dots, p_n \gets \textsc{SelectOrdering}(\calP)$
            \State $\vw^1 \gets 0, \; x^1 \gets \text{initial configuration}$
            \For{ $t = 1, \dots, T$ }
                \State $p^t \gets \textsc{SelectPart}(\calP)$
                \State $x^t_{p^t} \gets \argmax_{x_{p^t} \in \calX_{p^t}} u^t[J^t](x_{p^t} \circ x^{t}_{\overline{p}^t})$
                \State $x^t_{\overline{p}^t} \gets x^{t-1}_{\overline{p}^t}$
                \State User provides improvement $\hat{x}^t_{p^t}$ of $x^t_{p^t}$, keeping $x^t_{\overline{p}^t}$ fixed
                \IIf{ $\left(u^t[I^t](\hat{x}^t_{p^t} \circ x^t_{\overline{p}^t}) - u^t[I^t](x^t_{p^t} \circ x^t_{\overline{p}^t}) \le 0 \right)$ }
                    $\quad Q^t \gets I^t \quad $
                \IElse
                    $\quad Q^t \gets J^t$
                \EndIIf \label{eq:updatecondition}
                \State $\vw^{t+1}_{-Q^t} \gets \vw^t_{-Q^t}$
                \State $\vw^{t+1}_{ Q^t} \gets \vw^t_{ Q^t} + \vphi_{Q^t}(\hat{x}^t_{p^t} \circ x^t_{\overline{p}^t}) - \vphi_{Q^t}(x^t_{p^t} \circ x^t_{\overline{p}^t}))$
            \EndFor
        \EndProcedure
    \end{algorithmic}
\end{algorithm*}

\paragraph{The \pcl\ algorithm.} The pseudocode of our algorithm,
\pcl, is listed in Algorithm~\ref{alg:pcl}. \pcl\ starts off by
sorting the basic parts, producing an ordering $p_1, \ldots, p_n$.
Algorithm~\ref{alg:ordselect} could be employed or any other
(e.g. task-specific) sorting solution.  Then it loops for $T$
iterations, maintaining an estimate $\vw^t$ of the user weights as
well as a complete configuration $x^t$. The starting configuration
$x^1$ should be a reasonable initial guess, depending on the task. At
each iteration $t \in [T]$, the algorithm selects a part $p^t$ using
the procedure \textsc{SelectPart} (see below). Then it updates the
object $x^t$ by inferring a new partial configuration $x^t_{p^t}$
while keeping the rest of $x^t$ fixed, that is
$x^t_{\overline{p}^t} = x^{t-1}_{\overline{p}^t}$. The inferred partial
configuration $x^t_{p^t}$ is optimal with respect to the local
subutility $u^t_{I^t}(\cdot)$ given $x^{t-1}_{\overline{p}^t}$. Note that
inference is over the partial configuration $x^t_{p^t}$ only, and
therefore can be \emph{exponentially faster} than inference over full
configurations. 

Next, the algorithm presents the inferred partial configuration $x^t_{p^t}$ as
well as some contextual information (see below). The user is asked to produce an
improved partial configuration $\hat{x}^t_{p^t}$ according to the her own
preferences, while the rest of the object
$x^t_{\overline{p}^t}$ is kept fixed.  We assume that a user is satisfied with a
partial configuration $x^t_{p^t}$ if she cannot improve it further, or
equivalently when the object $x^t$ is \textit{conditionally optimal} with
respect to part $p^t$ given the rest of the object $x^t_{\overline{p}^t}$ (the
formal definition of conditional optimality is given in the Analysis section).
When a
user is satisfied with a partial configuration, she returns $\hat{x}^t_{p^t} =
x^t_{p^t}$, thereby implying no change in the weights $\vw^{t+1}$.

After receiving an improvement, if the user is not satisfied, the weights are
updated through a perceptron step. The subset $Q^t$ of weights that are actually
updated depends on whether $u^t[I^t](\hat{x}^t_{p^t} \circ x^t_{\overline{p}^t}) -
u^t[I^t](x^t_{p^t} \circ x^t_{\overline{p}^t})$ is negative or (strictly) positive.
Since we perform inference on $u^t[J^t](\cdot)$, we have that
$u^t[J^t](\overline{x}^t_{p^t} \circ x^t_{\overline{p}^t}) - u^t[J^t](\overline{x}^t_{p^t}
\circ x^t_{\overline{p}^t}) \le 0$. The user improvement can, however, potentially
change all the features in $I^t$. Intuitively, the weights associated to a
subset of features should change only if the utility computed on this subset
ranks $\hat{x}^t_{p^t}$ lower than $x^t_{p^t}$. The algorithm therefore checks
whether $u^t[I^t](\overline{x}^t_{p^t} \circ x^t_{\overline{p}^t}) \le u^t[I^t](x^t_{p^t}
\circ x^t_{\overline{p}^t})$, in which case the weights associated to the whole
subset $I^t$ should be updated. If this condition is not met, instead, the
algorithm can only safely update the weights associated to $J^t$, which, as
said, meet this condition by construction.

As for the \textsc{SelectPart} procedure, we experimented with several
alternative implementations, including prioritizing parts with a large feature
overlap ($|I_k \setminus J_k|$) and bandit-based strategies aimed at predicting
a surrogate of the utility gain (namely, a variant of the UCB1
algorithm~\cite{auer2002finite}). Preliminary experiments have shown
that informed strategies do not yield a significant performance
improvement over the random selection stategy; hence we stick with the
latter in all our experiments.

The algorithm stops either when the maximum number of iterations $T$ is met or
when a ``local optimum'' has been found. For ease of exposition we left out the
latter case from Algorithm~\ref{alg:pcl}, but we explain what a local optimum is
in the following Analysis section; the stopping criterion will follow directly
from Proposition~\ref{thm:locopt}.

\paragraph{Interacting through parts.} In order for the user to judge the
quality of a suggested partial configuration $x_p$, some contextual information
may have to be provided. The reason is that, if $p$ depends on other parts via
shared features, these have to be somehow communicated to the user, otherwise
his/her improvement will not be sufficiently informed.

We distinguish two cases, depending on whether the features of $p$ are local or
global. Local features only depend on small, localized portions of $x$. This is
for instance the case for features that measure the diversity of consecutive
activities in a touristic trip plan, which depend on consecutive time slots or
days only. Here the context amounts to those other portions of $x$ that
share local features with $p$. For instance, the user may interact over
individual days only. If the features are local, the context is simply the time
slots before and after the selected day. The user is free to modify the
activities scheduled that day based on the context, which is kept fixed.

On the other hand, global features depend on all of $x$ (or large
chunks of it). For instance, in house furnishing one may
have features that measure whether the total cost of the furniture is
within a given budget, or how much the cost surpasses the budget. A
naive solution would be that of showing the user the whole furniture
arrangement $x$, which can be troublesome when $x$ is large. A better
alternative is to present the user a \textit{summary} of the global
features, in this case the percentage of the used budget. Such a summary
would be sufficient for producing an informed improvement,
independently from the actual size of $x$.

Of course, the best choice of context format is application specific. We only
note that, while crucial, the context only provides auxiliary information
\textit{to the user}, and does not affect the learning algorithm directly.

\section{Analysis}

In preference elicitation, it is common to measure the quality of a recommended
(full) configuration in terms of the \emph{regret}:
$$ \textstyle \REG{x} := \max_{\hat{x}\in\calX} u^*(\hat{x})  - u^*(x) = \inner{\vw^*}{\vphi(x^*) - \vphi(x)} $$
where $u^*(\cdot)$ is the true, unobserved user utility and $x^* =
\argmax_{x\in\calX} u^*(x)$ is a truly optimal configuration.  In \pcl,
interaction with the user occurs via partial configurations, namely $x^t_{p^t}$
and $\hat{x}^t_{p^t}$. Since the regret is defined in terms of complete
configurations, it is difficult to analyze it directly based on information
about the partial configurations alone, making it hard to prove convergence to
globally optimal recommendations.

The aim of this analysis is, however, to show that our algorithm converges to a
locally optimal configuration, which is in line with guarantees offered by other
Coactive Learning variants~\cite{goetschalckx2014coactive}; the latter, however
still rely on interaction via complete configurations.  Here a configuration $x$
is a local optimum for $u^*(\cdot)$ if no part-wise modification can improve $x$
with respect to $u^*(\cdot)$. Formally, $x$ is a local optimum for $u^*(\cdot)$
if and only if:
\[ \forall \; p\!\in\!\calP \ \nexists \; x'_{p}\!\in\!\calX_{p} \quad u^*(x'_{p} \circ x_{\overline{p}}) > u^*(x_{p} \circ x_{\overline{p}}) \label{eq:localopt}\]
To measure local quality of a configuration $x$ with respect to a part $p$, we
introduce the concept of \emph{conditional regret} of the partial configuration
$x_{p}$ given the rest of the object $x_{\overline{p}}$:
$$ \textstyle \CREG{p}{x} := u^*(x^*_{p} \circ x_{\overline{p}}) - u^*(x_{p} \circ x_{\overline{p}}) $$
where $x^*_{p} = \argmax_{\hat{x}_{p}\in\calX_{p}} u^*(\hat{x}_{p} \circ
x_{\overline{p}})$. Notice that:
$$ \textstyle \CREG{p}{x} = u^*[I_p](x^*_{p} \circ x_{\overline{p}}) - u^*[I_p](x_{p} \circ x_{\overline{p}}) $$
since $u^*[-I_p](x^*_{p} \circ x_{\overline{p}}) - u^*[-I_p](x_{p} \circ
x_{\overline{p}}) = 0$.

We say that a partial configuration $x$ is \textit{conditionally optimal} with
respect to part $p$ if $\CREG{p}{x} = 0$. The following lemma gives sufficient
and necessary conditions for local optimality of a configuration $x$.

\begin{lem}
\label{lem:loisco}
A configuration $x$ is locally optimal with respect to $u^*(\cdot)$ if and only
if $x$ is conditionally optimal for $u^*(\cdot)$ with respect to all basic
parts $p\in \calP$.
\begin{proof}
By contradiction.
(i) Assume that $x$ is locally optimal but not conditionally optimal with
respect to $p \in \calP$. Then $\CREG{p}{x} > 0$, and thus there
exists a partial configuration $x'_p$ such that $u^*(x'_p \circ
x_{\bar{p}}) > u^*(x_p \circ x_{x_{\bar{p}}})$. This violates the
local optimality of $x$ (Eq.~4).
(ii) Assume that all partial configurations $x_p \ \forall p \in \calP$ are
conditionally optimal but $x$ is not locally optimal. Then there
exists a part $q \in \calP$ and a partial configuration $x'_q$ such that
$u^*(x'_q \circ x_{\bar{q}}) > u^*(x_q \circ x_{\bar{q}})$. This in turn
means that $\CREG{q}{x} > 0$. This violates the conditional optimality of
$x$ with respect to $q$.
\end{proof}
\end{lem}
%

The above lemma gives us a part-wise measurable criterion to determine if a
configuration $x$ is a local optimum through the conditional regret of $x$ for
all the provided parts.

The rest of the analysis is devoted to derive an upper bound on the conditional
regret incurred by the algorithm and to prove that \pcl\ eventually reaches a
local optimum.

In order to derive the bound, we rely on the concept of
$\alpha$-informativeness from~\cite{shivaswamy2015coactive}, adapting it to
part-wise interaction\footnote{Here we adopted the definition of
\textit{strict} $\alpha$-informativeness for simplicity. Our results
can be directly extended to the more general notions of informativeness described
in~\cite{shivaswamy2015coactive}.}.  A user is \emph{conditionally
$\alpha$-informative} if, when presented with a partial configuration
$x^t_{p^t}$, he/she provides a partial configuration $\hat{x}^t_{p^t}$ that is
at least some fraction $\alpha \in (0, 1]$ better than $x^t_{p^t}$ in terms of
conditional regret, or more formally:
\begin{align}
    & u^*[I^t](\hat{x}^t_{p^t} \circ x^t_{\overline{p}^t}) - u^*[I^t](x^t_{p^t} \circ x^t_{\overline{p}^t}) \ge \nonumber \\
    & \qquad\qquad \alpha \left(u^*[I^t](x^*_{p^t} \circ x^t_{\overline{p}^t}) - u^*[I^t](x^t_{p^t} \circ x^t_{\overline{p}^t})\right) \label{eq:ai}
\end{align}

\noindent
In the rest of the paper we will use the notation $u[I^t](\hat{x}^t)$ meaning
$u[I^t](\hat{x}^t_{p^t} \circ x^t_{\overline{p}^t})$, i.e. drop the complement, when
no ambiguity can arise.

At all iterations $t$, the algorithm updates the weights specified by $Q^t$,
producing a new estimate $\vw^{t+1}$ of $\vw^*$. The actual indices $Q^t$
depend on the condition at line~\ref{eq:updatecondition} of
Algorithm~\ref{alg:pcl}: at some iterations $Q^t$ includes all of $I^t$, while
at others $Q^t$ is restricted to $J^t$. We distinguish between these cases by:
\begin{align*}
    \calI &= \{ t \in [T] : u^t[I^t](\hat{x}^t) - u^t[I^t](x^t) \le 0 \} \\
    \calJ &= \{ t \in [T] : u^t[I^t](\hat{x}^t) - u^t[I^t](x^t) > 0 \}
\end{align*}
so that if $t\in\calI$ then $Q^t = I^t$, and $Q^t = J^t$ if $t\in\calJ$.
For all $t\in[T]$, the quality of $\vw^{t+1}$ is:
\begin{align*}
    \inner{\vw^*}{\vw^{t+1}} = \ & \inner{\vw^*_{-Q^t}}{\vw^{t+1}_{-Q^t}} + \inner{\vw^*_{Q^t}}{\vw^{t+1}_{Q^t}} \\
    = \ & \inner{\vw^*_{-Q^t}}{\vw^t_{-Q^t}} + \inner{\vw^*_{Q^t}}{\vw^t_{Q^t}} \\
        & \quad + \inner{\vw^*_{Q^t}}{\vphi_{Q^t}(\hat{x}^t) - \vphi_{Q^t}(x^t)} \\
    = \ & \inner{\vw^*}{\vw^t} + \inner{\vw^*_{Q^t}}{\vphi_{Q^t}(\hat{x}^t) - \vphi_{Q^t}(x^t)} \\
    = \ & \inner{\vw^*}{\vw^t} + u^*[Q^t](\hat{x}^t) - u^*[Q^t](x^t)
\end{align*}
Therefore if the second summand in the last equation, the \textit{utility gain}
$u^*[Q^t](\hat{x}^t) - u^*[Q^t](x^t)$, is positive, the update produces a better
weight estimate $\vw^{t+1}$.

Since the user is conditionally $\alpha$-informative, the improvement
$\hat{x}^t$ always satisfies $u^*[I^t](\hat{x}^t) - u^*[I^t](x^t) \ge 0$. When
$t\in\calI$, we have $Q^t = I^t$, and thus the utility gain is guaranteed to be
positive.
On the other hand, when $t\in\calJ$ we have $Q^t = J^t$ and the utility gain
reduces to $u^*[J^t](\hat{x}^t) - u^*[J^t](x^t)$. In this case the update
ignores the weights in $I^t \setminus J^t$, ``missing out'' a factor of
$u^*[I^t \setminus J^t](\hat{x}^t) - u^*[I^t \setminus J^t](x^t)$.

We compactly quantify the missing utility gain as:
$$
\zeta^t = \begin{cases}
    0 & \text{ if } t \in \calI \\
    u^*[I^t \setminus J^t](\hat{x}^t) - u^*[I^t \setminus J^t](x^t) & \text{ if } t \in \calJ
\end{cases}
$$
Note that $\zeta^t$ can be positive, null or negative for $t\in\calJ$. When
$\zeta^t$ is negative, making the update on $J^t$ only actually avoids a loss in
utility gain.

We now prove that \pcl\ minimizes the average conditional regret
$\ACREG{T} := \frac{1}{T} \sum_{t=1}^T \CREG{p^t}{x^t}$ as $T \to \infty$
for conditionally $\alpha$-informative users.
\begin{thm}
\label{thm:bound}
For a conditionally $\alpha$-informative user, the average conditional regret of
\pcl\ after $T$ iterations is upper bounded by:
$$ \frac{1}{T} \sum_{t=1}^T \CREG{p^t}{x^t} \le \frac{2DS\norm{\vw^*}}{\alpha \sqrt{T}} + \frac{1}{\alpha T} \sum_{t=1}^T \zeta^t $$
\begin{proof}
The following is a sketch\footnote{The complete proof can be found in the
Supplementary Material.}.
We start by splitting the iterations into the $\calI$ and $\calJ$ sets defined
above, and bound the norm $\norm{\vw^{T+1}}^2$. In both cases we find that
$\norm{\vw^{T+1}}^2 \le 4D^2S^2T$. We then expand the term
$\inner{\vw^*}{\vw^{T+1}}$ for iterations in both $\calI$ and $\calJ$,
obtaining:
\begin{align*}
    \inner{\vw^*}{\vw^{T+1}} &= \sum_{t\in\calI} \inner{\vw^*_{I^t}}{\vphi_{I^t}(\hat{x}^t) - \vphi_{I^t}(x^t)} \\
                             &\qquad + \sum_{t\in\calJ} \inner{\vw^*_{J^t}}{\vphi_{J^t}(\hat{x}^t) - \vphi_{J^t}(x^t)}
\end{align*}
With few algebraic manipulations we obtain:
\begin{align*}
          & \inner{\vw^*}{\vw^{T+1}} \\
      = \ & \sum_{t=1}^T \inner{\vw^*_{I^t}}{\vphi_{I^t}(\hat{x}^t) - \vphi_{I^t}(x^t)} - \sum_{t=1}^T \zeta^t \\
\end{align*}
Which we then bound using the Cauchy-Schwarz inequality:
\begin{align*}
          & \norm{\vw^*}\sqrt{4D^2S^2T} \\
    \ge \ & \norm{\vw^*}\norm{\vw^{T+1}} \\
    \ge \ & \inner{\vw^*}{\vw^{T+1}} \\
      = \ & \sum_{t=1}^T \inner{\vw^*_{I^t}}{\vphi_{I^t}(\hat{x}^t) - \vphi_{I^t}(x^t)} - \sum_{t=1}^T \zeta^t \\
\end{align*}
Applying the conditional $\alpha$-informative feedback (Eq.~\ref{eq:ai}) and
rearranging proves the claim.
\end{proof}
\end{thm}

Theorem~\ref{thm:bound} ensures that the average conditional regret suffered
by our algorithm decreases as $\calO(1/\sqrt{T})$. This alone, however, does not
prove that the algorithm will eventually arrive at a local optimum, even if
$\sum_{t=\tau_0}^T \CREG{p^t}{x^t} = 0$, for some $\tau_0\in[T]$. This is due to
the fact that partial inference is performed keeping the rest of the object
$x^t_{\overline{p}^t}$ fixed. Between iterations an inferred part may change as a
result of a change of the other parts in previous iterations. The object could,
in principle, keep changing at every iterations, even if $\CREG{p^t}{x^t}$ is
always equal to $0$. The next proposition, however, shows that this is not the
case thanks to the utility decomposition we employ.
\begin{prop}
\label{thm:locopt}
Let $\tau_{1} \!<\! \dots \!<\! \tau_{n} \!<\! \hat{\tau}_{1} \!<\! \dots \!<\!
\hat{\tau}_{n} \!\le\! T$ such that $p^{\tau_{k}} = p^{\hat{\tau}_k} = p_k$ and
$\CREG{p^t}{x^t} = 0$ for all $t \ge \tau_{1}$. The configuration $x^T$ is a
local optimum.
\begin{proof}
Sketch.
The proof procedes by strong induction. We first show that $x^t_{p_1} =
x^{\tau_1}_{p_1}$ for all $t > \tau_1$, as $u^t_{I_1}(\cdot)$ only depends on
the features in $J_1$ and by assumption $\CREG{p^t}{x^t} = 0$ for all $t \ge
\tau_1$.
By strong induction, assuming that $x^t_{p_j} = x^{\tau_j}_{p_j}$ for all $j =
1, \dots, k-1$ and all $t > \tau_{k-1}$, we can easily show that $x^t_{p_k} =
x^{\tau_k}_{p_k}$ as well.

Now, $x^t_{p_k} = x^{\tau_k}_{p_k}$ for all $t>\tau_n$, therefore
$x^{\hat{\tau}_k} = x^{\tau_n}$ for all $k\in[n]$. Since
$\CREG{p_k}{x^{\hat{\tau}_k}} = 0$ for all $k\in[n]$ by assumption, then $x^T =
x^{\hat{\tau}_n} = x^{\tau_n}$ is a local optimum and will not change for all
$t>\tau_n$.

\end{proof}
\end{prop}

The algorithm actually reaches a local optimum at $t = \tau_n$, but it needs to
double check all the parts in order to be sure that the configuration is
actually a local optimum. This justifies a termination criterion that we use in
practice: if the algorithm completes two full runs over all the parts, and the
user can never improve any of the recommended partial configurations, then the
full configuration is guaranteed to be a local optimum, and the algorithm can
stop. As mentioned, we employ this criterion in our implementation but we left
it out from Algorithm~\ref{alg:pcl} for simplicity.

To recap, Theorem~\ref{thm:bound} guarantees that $\ACREG{T} \to 0$ as
$t\to\infty$, therefore $\CREG{p_t}{x^t}$ approaches $0$ as well. Combining
this fact with Proposition~\ref{thm:locopt}, we proved that our algorithm
approaches a local optimum for $T\to\infty$.

\section{Empirical Analysis}

We ran \pcl\ on three constructive preference elicitation tasks of increasing
complexity, comparing different degrees of user informativeness. According to
our experiments, informativeness is the most critical factor. The three
problems involve rather large configurations, which can not be handled by
coactive interaction via complete configurations. For instance, in
\cite{ortega2016human} the user is tasked to solve relatively simple SAT
instances over three variables and (at most) eight clauses; in some cases users
were observed to show signs of cognitive overload. In comparison, our simplest
realistic problem involve 35 categorical variables (with 8 possible values) and
74 features, plus additional hard constraints. As a consequence, Coactive
Learning can not be applied as-is, and part-wise interaction is necessary.

In all of these settings, part-wise inference is cast as a mixed integer linear
problem (MILP), and solved with
Gecode\footnote{\texttt{http://www.gecode.org/}}.  Despite being NP-hard in
general, MILP solvers can be very efficient on practical instances. Efficiency
is further improved by inferring only partial configurations. Our experimental
setup is available at \texttt{https://github.com/unitn-sml/pcl}.

We employed a user simulation protocol similar to that
of~\cite{teso2017coactive}. First, for each problem, we sampled $20$
vectors $\vw^*$ at random from a standard normal distribution. Then,
upon receiving a recommendation
$x^t_{p^t}$, 
an improvement $\hat{x}^t_{p^t}$ is generated by solving the following
problem:
\begin{align*}
    \argmin_{x_{p^t} \in \calX_{p^t}} \ \ & u^*[I^t](x_{p^t} \circ x^t_{\overline{p}^t}) \\
    \text{s.t.} \quad
        & u^*[I^t](x_{p^t} \circ x^t_{\overline{p}^t}) - u^*[I^t](x^t_{p^t} \circ x^t_{\overline{p}^t}) \\
        & \qquad \ge \alpha (u^*[I^t](x^*_{p^t} \circ x^t_{\overline{p}^t}) - u^*[I^t](x^t_{p^t} \circ x^t_{\overline{p}^t}))
\end{align*}
This formulation clearly satisfies the conditional $\alpha$-informativeness
assumption (Eq.~\ref{eq:ai}).

\begin{figure}[h]
    \centering
    \includegraphics[width=0.25\textwidth]{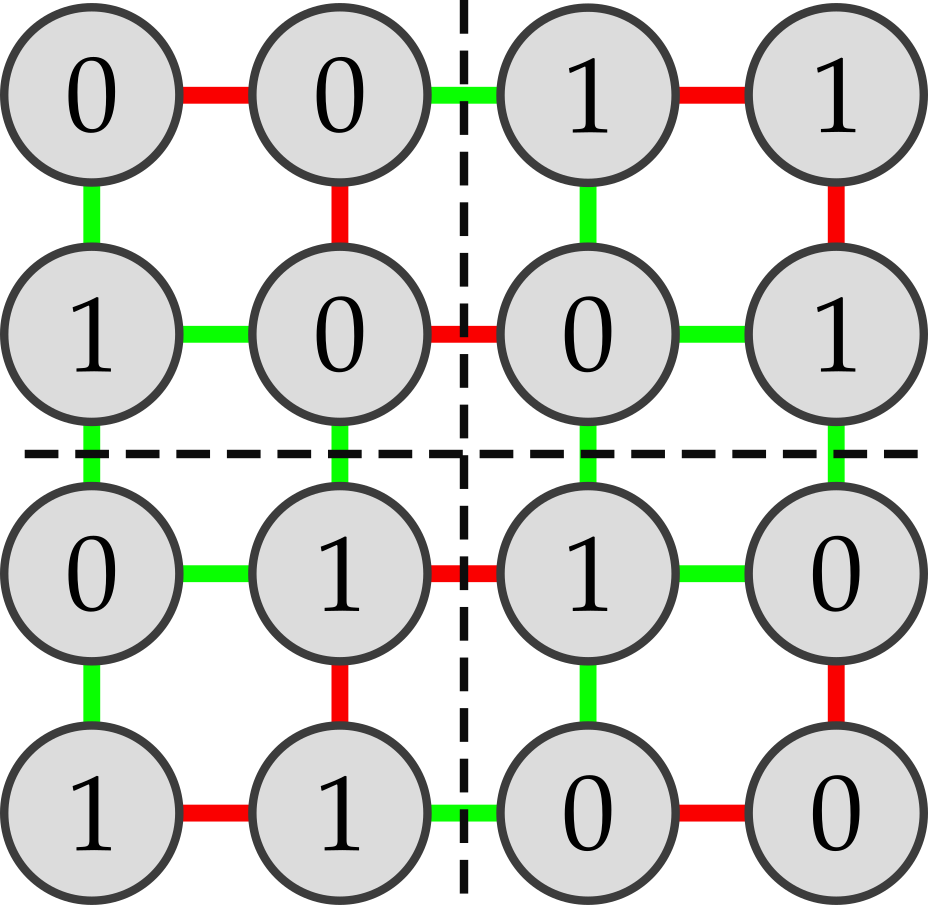}
    \caption{\label{fig:graph} Example grid configuration.}
\end{figure}

\paragraph{Synthetic setting.} We designed a simple synthetic
problem inspired by spin glass models, see Figure~\ref{fig:graph} for a
depiction. In this setting, a configuration $x$
consists of a $4 \times 4$ grid. Each node in the grid is a binary 0-1
variable. Adjacent nodes are connected by an edge, and each edge is
associated to an indicator feature that evaluates to $1$ if the
incident nodes have different values (green in the figure), and to $-1$ otherwise
(red in the figure). The utility of a
configuration is simply the weighted sum of the values of all features (edges).
The basic parts $p$ consist of all the non-overlapping $2 \times 2$ sub-grids
of $x$, for a total of $4$ basic parts (indicated by dotted lines in the
figure).

Since the problem is small enough for inference of complete configurations to
be practical, we compared \pcl\ to standard Coactive Learning, using the
implementation of~\cite{teso2017coactive}. In order to keep the comparison as
fair as possible, the improvements fed to CL were chosen to match the utility
gain obtained by \pcl. We further report the performance of three alternative
part selection strategies: random, smallest (most independent) part first, and
UCB1.

\begin{figure*}[t]
    \centering
    \begin{tabular}{ccc}
        \includegraphics[width=0.3\textwidth]{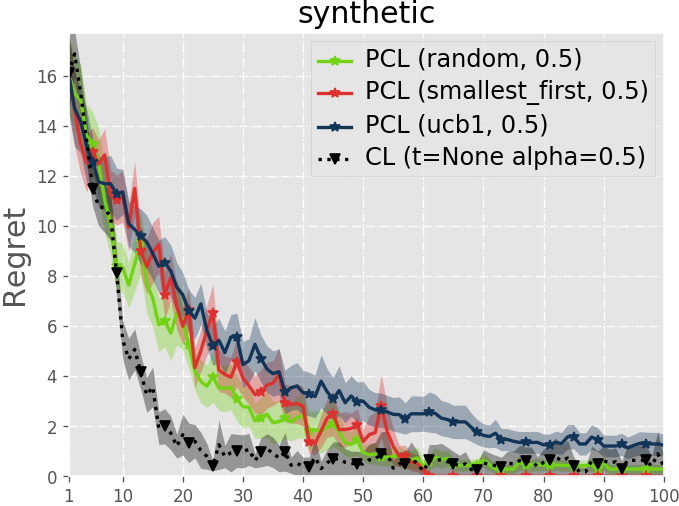} &
        \includegraphics[width=0.3\textwidth]{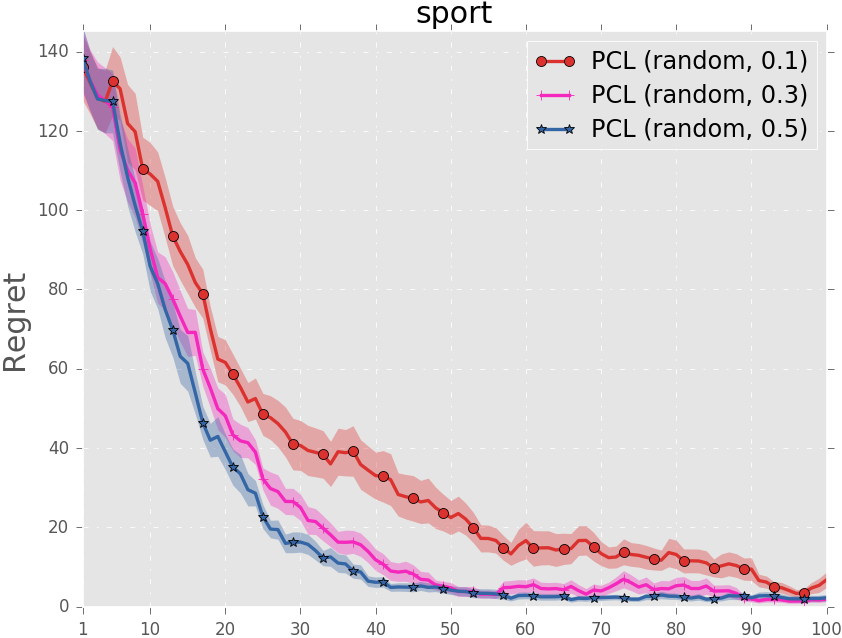} &
        \includegraphics[width=0.3\textwidth]{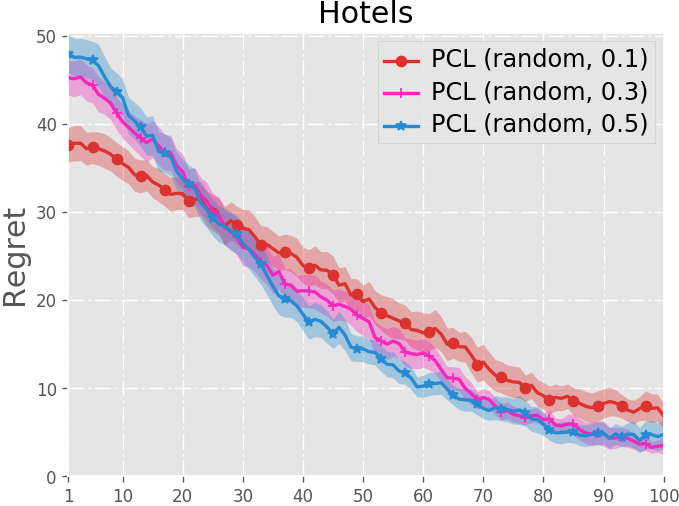}
        \\
        \includegraphics[width=0.3\textwidth]{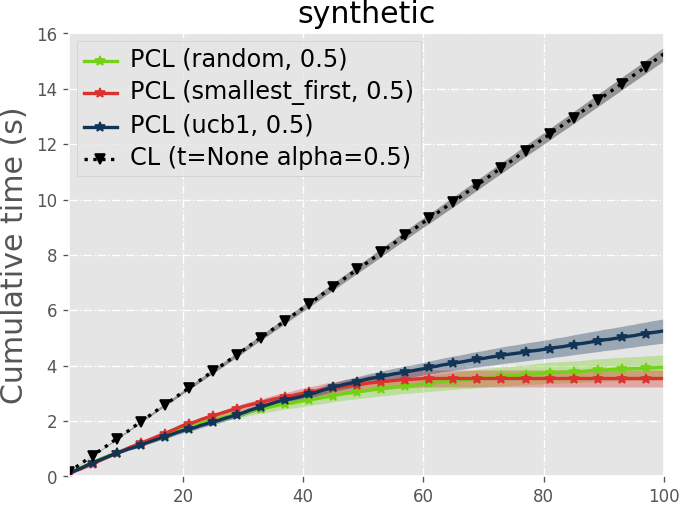} &
        \includegraphics[width=0.3\textwidth]{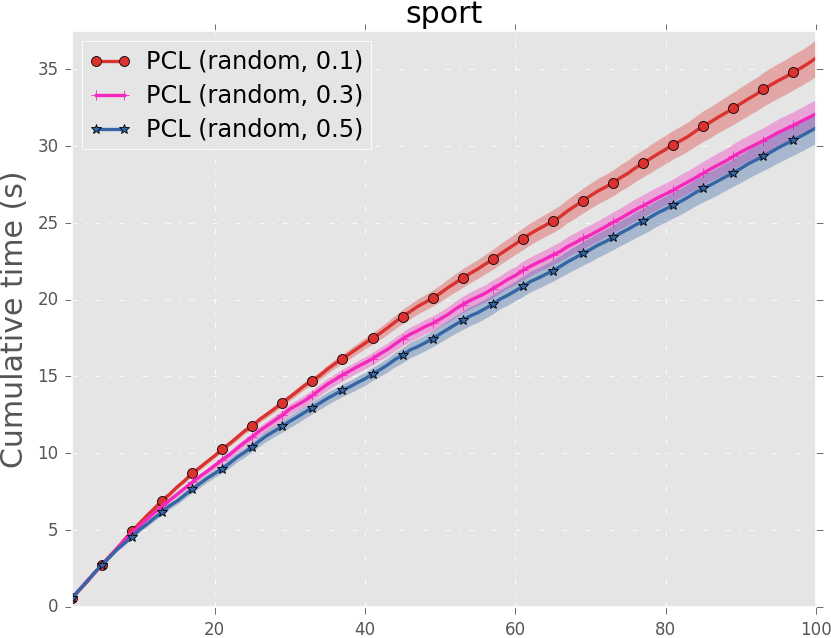} &
        \includegraphics[width=0.3\textwidth]{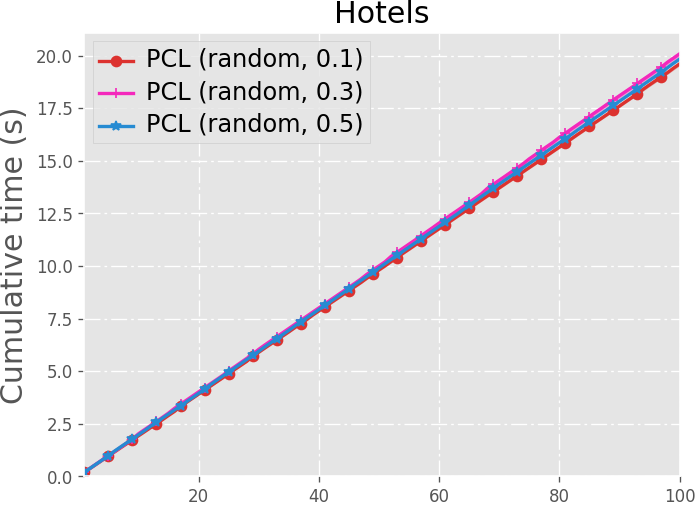}
    \end{tabular}
    \caption{\label{fig:results} Regret over \emph{complete} configurations (top) and cumulative runtime (bottom)
    of \pcl\ and CL on our three constructive problems: synthetic (left), training
    planning (middle), and hotel planning (right). The $x$-axis is the number
    of iterations, while the shaded areas represent the standard deviation.
    Best viewed in color.}
\end{figure*}

The results can be found in the first column of
Figure~\ref{fig:results}. We report both the regret (over \emph{complete}
configurations) and the cumulative
runtime of all algorithms, averaged over all users, as well as their
standard deviation. The regret plot shows that, despite being
restricted to work with $2 \times 2$ configurations, \pcl\ does
recommend \emph{complete} configurations of quality comparable to CL
after enough queries are made.  Out of the three part selection
strategies, random performs best, with the other two more informed
alternatives (especially smallest first) quite close.
The runtime gap between full and part-wise inference is already clear in this
small synthetic problem; complete inference quickly becomes impractical as the
problem size increases.

\paragraph{Training planning.} Generating personalized training plans based on
performance and health monitoring has received a lot of attention recently in
sport analytics (see e.g.~\cite{fister2015planning}).
Here we consider the problem of synthesizing a week-long training plan $x$ from
information about the target athlete. Each day includes 5 time slots (two for
the morning, two for the afternoon, one for the evening), for $35$ slots total.
We assume to be given a fixed number of training activities ($7$ in our
experiments: walking, running, swimming, weight lifting, pushups, squats, abs),
as well as knowledge of
the slots in which the athlete is available. The training plan $x$ associates
an activity to each slot where the athlete is available. Our formulation tracks
the amount of improvement (e.g. power increase) and fatigue over five different
body parts (arms, torso, back, legs, and heart) induced by performing an
activity for one time slot. Each day defines a basic part.

The mapping between training activity and improvement/fatigue over each body
part is assumed to be provided externally. It can be provided by the athlete
or medical personnel monitoring his/her status. The features of $x$ include, for
each body part, the total performance gain and fatigue, computed over the
recommended training plan according to the aforementioned mapping. We further
include inter-part features to capture activity diversity in consecutive days.
The fatigue accumulated in 3 consecutive time slots in any body parts does not
exceed a given threshold, to prevent injuries.

In this setting, CL is impractical from both the cognitive and computational
points of view. We ran \pcl\ and evaluated the impact of user informativeness
by progressively increasing $\alpha$ from $0.1$, to $0.3$, to $0.5$.
The results can be seen in Figure~\ref{fig:results}. The plots show clearly
that, despite the complexity of the configuration and constraints, \pcl\ can
still produce very low-regret configurations after about 50 iterations or less.

Understandably, the degree of improvement $\alpha$ plays an important role in
the performance of \pcl\ and, consequently, in its runtime (users at
convergence do not contribute to the runtime), at least up to $\alpha=0.5$.
Recall, however, that the improvements are part-wise, and hence $\alpha$
quantifies the degree of \emph{local} improvement: part improvements may be
very informative on their own, but only give a modest amount of information
about the full configuration.  However, it is not unreasonable to expect that
users to be very informative when presented with reasonably sized (and simple)
parts. Crucially \pcl\ allows the system designer to define the parts
appropriately depending on the application.

\paragraph{Hotel planning.}
Finally, we considered a complex furniture allocation problem:
furnishing an entire hotel. The problem is encoded as follows. The
hotel is represented by a graph: nodes are rooms and edges indicate
which rooms are adjacent. Rooms can be of three types: normal rooms,
suites, and dorms. Each room can hold a maximum number of furniture
pieces, each associated to a cost. Additional, fixed nodes represent
bathrooms and bars. The type of a room is decided dynamically based on
its position and furniture. For instance, a normal room must contain
at most three single or double beds, no bunk beds, and a table, and
must be close to a bathroom. A suite must contain one bed, a table and
a sofa, and must be close to a bathroom and a bar. Each room is a
basic part, and there are 15 rooms to be allocated.

The feature vector contains $20$ global features plus $8$ local features per
room.  The global features include different functions of the number of
different types of rooms, the total cost of the furniture and the total number
of guests. The local features include, instead, characteristics of the current
room, such as its type or the amount of furniture, and other features shared by
adjacent rooms, e.g. whether two rooms have the same type.  These can encode
preferences like ``suites and dorms should not be too close'', or ``the hotel
should maintain high quality standards while still being profitable''.
Given the graph structure, room capacities, and total budget, the goal is to
furnish all rooms according to the user's preferences.

This problem is hard to solve to optimality with current solvers;
part-based inference alleviates this issue by focusing on individual
rooms. There are 15 rooms in the hotel, so that at each iteration only
1/15 of the configuration is affected. Furthermore, the presence
of the global features implies dependences between all rooms.
Nonetheless, the algorithm manages to reduce the regret by an order of
magnitude in around a 100 iterations, starting from a completely
uninformed prior. Note also that as for the training planning
scenario, an alpha of 0.3 achieves basically the same results as those
for alpha equal to 0.5.

\section{Conclusion}

In this work we presented an approach to constructive preference elicitation
able to tackle large constructive domains, beyond the reach of previous
approaches. It is based on Coactive Learning~\cite{shivaswamy2015coactive}, but
only requires inference of partial configurations and partial improvement
feedback, thereby significantly reducing the cognitive load of the user. We
presented an extensive theoretical analysis demonstrating that, despite working
only with partial configurations, the algorithm converges to a locally optimal
solution.  The algorithm has been evaluated empirically on three constructive
scenarios of increasing complexity, and shown to perform well in practice.

Possible future work includes improving part-based interaction by exchanging
additional contextual information (e.g. features~\cite{teso2017coactive} or
explanations) with the user, and applying \pcl\ to large layout synthesis
problems~\cite{dragone2016layout}.

\bibliographystyle{aaai}
\bibliography{paper}

\begin{thebibliography}{}

\bibitem[\protect\citeauthoryear{Auer, Cesa-Bianchi, and
  Fischer}{2002}]{auer2002finite}
Auer, P.; Cesa-Bianchi, N.; and Fischer, P.
\newblock 2002.
\newblock Finite-time analysis of the multiarmed bandit problem.
\newblock {\em Machine learning} 47(2-3):235--256.

\bibitem[\protect\citeauthoryear{Boutilier, Bacchus, and
  Brafman}{2001}]{boutilier2001ucp}
Boutilier, C.; Bacchus, F.; and Brafman, R.~I.
\newblock 2001.
\newblock Ucp-networks: A directed graphical representation of conditional
  utilities.
\newblock In {\em Proceedings of the Seventeenth conference on Uncertainty in
  artificial intelligence},  56--64.
\newblock Morgan Kaufmann Publishers Inc.

\bibitem[\protect\citeauthoryear{Boutilier \bgroup et al\mbox.\egroup
  }{2006}]{boutilier2006constraint}
Boutilier, C.; Patrascu, R.; Poupart, P.; and Schuurmans, D.
\newblock 2006.
\newblock Constraint-based optimization and utility elicitation using the
  minimax decision criterion.
\newblock {\em Artificial Intelligence} 170(8-9):686--713.

\bibitem[\protect\citeauthoryear{Braziunas and
  Boutilier}{2005}]{braziunas2005local}
Braziunas, D., and Boutilier, C.
\newblock 2005.
\newblock Local utility elicitation in {GAI} models.
\newblock In {\em Proceedings of the Twenty-First Conference on Uncertainty in
  Artificial Intelligence},  42--49.
\newblock AUAI Press.

\bibitem[\protect\citeauthoryear{Braziunas and
  Boutilier}{2007}]{braziunas2007minimax}
Braziunas, D., and Boutilier, C.
\newblock 2007.
\newblock Minimax regret based elicitation of generalized additive utilities.
\newblock In {\em UAI},  25--32.

\bibitem[\protect\citeauthoryear{Braziunas and
  Boutilier}{2009}]{braziunas2009elicitation}
Braziunas, D., and Boutilier, C.
\newblock 2009.
\newblock Elicitation of factored utilities.
\newblock {\em AI Magazine} 29(4):79.

\bibitem[\protect\citeauthoryear{Chajewska, Koller, and
  Parr}{2000}]{chajewska2000making}
Chajewska, U.; Koller, D.; and Parr, R.
\newblock 2000.
\newblock Making rational decisions using adaptive utility elicitation.
\newblock In {\em AAAI/IAAI},  363--369.

\bibitem[\protect\citeauthoryear{Dragone \bgroup et al\mbox.\egroup
  }{2016}]{dragone2016layout}
Dragone, P.; Erculiani, L.; Chietera, M.~T.; Teso, S.; and Passerini, A.
\newblock 2016.
\newblock Constructive layout synthesis via coactive learning.
\newblock In {\em Constructive Machine Learning workshop, NIPS}.

\bibitem[\protect\citeauthoryear{Fishburn}{1967}]{fishburn1967interdependence}
Fishburn, P.~C.
\newblock 1967.
\newblock Interdependence and additivity in multivariate, unidimensional
  expected utility theory.
\newblock {\em International Economic Review} 8(3):335--342.

\bibitem[\protect\citeauthoryear{Fister \bgroup et al\mbox.\egroup
  }{2015}]{fister2015planning}
Fister, I.; Rauter, S.; Yang, X.-S.; and Ljubi{\v{c}}, K.
\newblock 2015.
\newblock Planning the sports training sessions with the bat algorithm.
\newblock {\em Neurocomputing} 149:993--1002.

\bibitem[\protect\citeauthoryear{Goetschalckx, Fern, and
  Tadepalli}{2014}]{goetschalckx2014coactive}
Goetschalckx, R.; Fern, A.; and Tadepalli, P.
\newblock 2014.
\newblock Coactive learning for locally optimal problem solving.
\newblock In {\em Proceedings of AAAI}.

\bibitem[\protect\citeauthoryear{Gonzales and Perny}{2004}]{gonzales2004gai}
Gonzales, C., and Perny, P.
\newblock 2004.
\newblock {GAI} networks for utility elicitation.
\newblock {\em KR} 4:224--234.

\bibitem[\protect\citeauthoryear{Keeney and Raiffa}{1976}]{keeney1976}
Keeney, R.~L., and Raiffa, H.
\newblock 1976.
\newblock {\em Decisions with Multiple Objectives: Preferences and Value
  Tradeoffs}.

\bibitem[\protect\citeauthoryear{Mayer and Moreno}{2003}]{mayer2003nine}
Mayer, R.~E., and Moreno, R.
\newblock 2003.
\newblock Nine ways to reduce cognitive load in multimedia learning.
\newblock {\em Educational psychologist} 38(1):43--52.

\bibitem[\protect\citeauthoryear{Meseguer, Rossi, and
  Schiex}{2006}]{meseguer2006soft}
Meseguer, P.; Rossi, F.; and Schiex, T.
\newblock 2006.
\newblock Soft constraints.
\newblock {\em Foundations of Artificial Intelligence} 2:281--328.

\bibitem[\protect\citeauthoryear{Ortega and Stocker}{2016}]{ortega2016human}
Ortega, P.~A., and Stocker, A.~A.
\newblock 2016.
\newblock Human decision-making under limited time.
\newblock In {\em Advances in Neural Information Processing Systems},
  100--108.

\bibitem[\protect\citeauthoryear{Pigozzi, Tsouki{\`{a}}s, and
  Viappiani}{2016}]{pigozzi16preferences}
Pigozzi, G.; Tsouki{\`{a}}s, A.; and Viappiani, P.
\newblock 2016.
\newblock Preferences in artificial intelligence.
\newblock {\em Ann. Math. Artif. Intell.} 77(3-4):361--401.

\bibitem[\protect\citeauthoryear{Raman, Shivaswamy, and
  Joachims}{2012}]{raman2012online}
Raman, K.; Shivaswamy, P.; and Joachims, T.
\newblock 2012.
\newblock Online learning to diversify from implicit feedback.
\newblock In {\em Proceedings of the 18th ACM SIGKDD international conference
  on Knowledge discovery and data mining},  705--713.
\newblock ACM.

\bibitem[\protect\citeauthoryear{Shivaswamy and
  Joachims}{2015}]{shivaswamy2015coactive}
Shivaswamy, P., and Joachims, T.
\newblock 2015.
\newblock Coactive learning.
\newblock {\em JAIR} 53:1--40.

\bibitem[\protect\citeauthoryear{Teso, Dragone, and
  Passerini}{2017}]{teso2017coactive}
Teso, S.; Dragone, P.; and Passerini, A.
\newblock 2017.
\newblock Coactive critiquing: Elicitation of preferences and features.
\newblock In {\em AAAI}.

\bibitem[\protect\citeauthoryear{Teso, Passerini, and
  Viappiani}{2016}]{teso2016constructive}
Teso, S.; Passerini, A.; and Viappiani, P.
\newblock 2016.
\newblock Constructive preference elicitation by setwise max-margin learning.
\newblock In {\em Proceedings of the Twenty-Fifth International Joint
  Conference on Artificial Intelligence},  2067--2073.
\newblock AAAI Press.

\end{thebibliography}

\newpage
\appendix

\section{Supplementary Material}

\subsection{Proof of Theorem~2}

We begin by expanding $\norm{\vw^{T+1}}^2$. If $T \in \calI$:
\begin{align*}
          & \norm{\vw^{T+1}}^2 \\
      = \ & \norm{\vw^{T+1}_{-I^T}}^2 + \norm{\vw^{T+1}_{I^T}}^2 \\
      = \ & \norm{\vw^{T}_{-I^T}}^2 + \norm{\vw^T_{I^T} + \vphi_{I^T}(\hat{x}^t) - \vphi_{I^T}(x^t)}^2 \\
      = \ & \norm{\vw^{T}_{-I^T}}^2 + \norm{\vw^T_{I^T}}^2 + \norm{\vphi_{I^T}(\hat{x}^t) - \vphi_{I^T}(x^t)}^2 \\
          & \qquad + 2 \inner{\vw^T_{I^T}}{\vphi_{I^T}(\hat{x}^t) - \vphi_{I^T}(x^t)}
\end{align*}
Since $T\in\calI$, $u^T[I^T](\hat{x}^T) - u^T[I^T](x^T) \le 0$, thus:
\begin{align*}
          & \norm{\vw^{T+1}}^2 \\
    \le \ & \norm{\vw^{T}_{-I^T}}^2 + \norm{\vw^T_{I^T}}^2 + \norm{\vphi_{I^T}(\hat{x}^t) - \vphi_{I^T}(x^t)}^2_{\infty} |I^T|^2 \\
    \le \ & \norm{\vw^T}^2 + 4 D^2 S^2
\end{align*}
If instead $T\in\calJ$:
\begin{align*}
          & \norm{\vw^{T+1}}^2 \\
      = \ & \norm{\vw^{T+1}_{-J^T}}^2 + \norm{\vw^{T+1}_{J^T}}^2 \\
      = \ & \norm{\vw^{T}_{-J^T}}^2 + \norm{\vw^{T+1}_{J^T}}^2 \\
      = \ & \norm{\vw^{T}_{-J^T}}^2 + \norm{\vw^T_{J^T} + \vphi_{J^T}(\hat{x}^t) - \vphi_{J^T}(x^t)}^2 \\
      = \ & \norm{\vw^{T}_{-J^T}}^2 + \norm{\vw^T_{J^T}}^2 + \norm{\vphi_{J^T}(\hat{x}^t) - \vphi_{J^T}(x^t)}^2 \\
          & \qquad + 2\inner{\vw^T_{J^T}}{\vphi_{J^T}(\hat{x}^t) - \vphi_{J^T}(x^t)} \\
    \le \ & \norm{\vw^{T}}^2 + \norm{\vphi_{J^T}(\hat{x}^t) - \vphi_{J^T}(x^t)}^2_{\infty} |J^T|^2 \\
    \le \ & \norm{\vw^T}^2 + 4 D^2 |J^T|^2 \\
    \le \ & \norm{\vw^T}^2 + 4 D^2 S^2
\end{align*}
We can therefore expand the term:
\begin{align*}
    \norm{\vw^{T+1}}^2 &\le 4 D^2 S^2 T
\end{align*}

Applying Cauchy-Schwarz inequality:
\begin{align*}
    \inner{\vw^*}{\vw^{T+1}} &\le \norm{\vw^*}\norm{\vw^{T+1}} \\
                             &\le 2 D S \norm{\vw^*} \sqrt{T}
\end{align*}
The LHS of the above inequality expands to:
\begin{align*}
    \inner{\vw^*}{\vw^{T+1}} &= \inner{\vw^*_{-I^T}}{\vw^T_{-I^T}} + \inner{\vw^*_{I^T}}{\vw^T_{I^T}} \\
                             & \qquad + \inner{\vw^*_{I^T}}{\vphi_{I^T}(\hat{x}^T) - \vphi_{I^T}(x^T)} \\
    [\text{ if } T \in \calI \ ] \\
    \inner{\vw^*}{\vw^{T+1}} &= \inner{\vw^*_{-J^T}}{\vw^T_{-J^T}} + \inner{\vw^*_{J^T}}{\vw^T_{J^T}} \\
                             & \qquad + \inner{\vw^*_{J^T}}{\vphi_{J^T}(\hat{x}^T) - \vphi_{J^T}(x^T)} \\
    [\text{ if } T \in \calJ \ ]
\end{align*}
And thus:
\begin{align*}
    \inner{\vw^*}{\vw^{T+1}} &= \sum_{t\in\calI} \inner{\vw^*_{I^t}}{\vphi_{I^t}(\hat{x}^t) - \vphi_{I^t}(x^t)} \\
                             &\qquad + \sum_{t\in\calJ} \inner{\vw^*_{J^t}}{\vphi_{J^t}(\hat{x}^t) - \vphi_{J^t}(x^t)}
\end{align*}
We add and subtract the term:
$$ \sum_{t\in\calJ} \inner{\vw^*_{I^t \setminus J^t}}{\vphi_{I^t \setminus J^t}(\hat{x}^t) - \vphi_{I^t \setminus J^t}(x^t)} $$
We obtain:
\begin{align*}
          & \inner{\vw^*}{\vw^{T+1}} \\
      = \ & \sum_{t=1}^T \inner{\vw^*_{I^t}}{\vphi_{I^t}(\hat{x}^t) - \vphi_{I^t}(x^t)} \\
                             &\qquad - \sum_{t\in\calJ} \inner{\vw^*_{I^t \setminus J^t}}{\vphi_{I^t \setminus J^t}(\hat{x}^t) - \vphi_{J^t}(x^t)} \\
      = \ & \sum_{t=1}^T \inner{\vw^*_{I^t}}{\vphi_{I^t}(\hat{x}^t) - \vphi_{I^t}(x^t)} - \sum_{t=1}^T \zeta^t \\
    \ge \ & \alpha \sum_{t=1}^T \inner{\vw^*_{I^t}}{\vphi_{I^t}(x^*_{p^t} \circ x^t_{\bar{p^t}}) - \vphi_{I^t}(x^t)} - \sum_{t=1}^T \zeta^t \\
\end{align*}
And thus:
$$ \sum_{t=1}^T \inner{\vw^*_{I^t}}{\vphi_{I^t}(x^*) - \vphi_{I^t}(x^t)} \le \frac{2 D S \norm{\vw^*} \sqrt{T}}{\alpha} + \frac{1}{\alpha} \sum_{t=1}^T \zeta^t $$
Dividing both sides by $T$ proves the claim.

\subsection{Proof of Proposition~3}

We first show that $x^t_{p_k} = x^{\tau_k}_{p_k}$ for $t > \tau_k$. Recall that
at iteration $t$ the algorithm produces $x^{t}_{p_k}$ by maximizing the local
subutility:
\begin{align*}
\textstyle u^t_{I_k}(x) &= \textstyle u^t[J_k](x) = \textstyle u^t[I_k \setminus ( \bigcup_{j=k+1}^n I_j )](x)
\end{align*}
i.e. ignoring all the features in $\bigcup_{j=k+1}^{n} I_j$. Each subutility
$u^t_{I_k}(\cdot)$ is only affected by the changes in $x_{p_1}, \dots,
x_{p_{k-1}}$, as their feature subsets may intersect with $I_k$.

The subutility $u^{t}_{I_1}(\cdot)$ only depends on the features that are
exclusively included in $I_1$, so it is not affected by the changes in all the
other parts. At iteration $\tau_1$, the conditional regret
$\CREG{p_1}{x^{\tau_1}} = 0$, and thus $\hat{x}^{\tau_1}_{p_1} =
x^{\tau_1}_{p_1}$ and $\vw^{\tau_1 + 1}_{I^t} = \vw^{\tau_1}_{I^t}$. If $p_1$
gets selected again at $t > \tau_1$, $u^t_{I_1}(\cdot)$ will not have changed as
it only depends on features exclusively included in $I_1$ and thus $x^t_{p_1} =
x^{\tau_1}_{p_1}$. Since, by assumption, $\CREG{p^t}{x^t} = 0$ for all $t \ge
\tau_1$, $u^{\tau_1}_{I_1}(\cdot)$ will not change anymore and so $x^t_{p_1} = x^{\tau_1}_{p_1}$.

By strong induction, suppose that all parts $x^t_{p_j} = x^{\tau_{j}}_{p_{j}}$ for all
$j = 1, \dots k-1$ and for all $t > \tau_{k-1}$. At iteration $\tau_k$,
$\CREG{p_k}{x^{\tau_k}} = 0$, thus $\hat{x}^{\tau_k}_{p_k} = x^{\tau_k}_{p_k}$
and so the weights $\vw^{\tau_k + 1}_{I_k}$ will not change.
For the
following $t > \tau_k$ iterations, even if $p_k$ gets selected, since the
subutility $u^t_{p_k}(\cdot)$ is only affected by the changes in $x_{p_{1}},
\dots, x_{p_{k-1}}$ and those parts do not change (by inductive assumption), we
can conclude that $u^t_{p_k}(\cdot) = u^{\tau_k}_{p_k}(\cdot)$ and that
$x^t_{p_k} = x^{\tau_k}_{p_k}$.

This proves that, for any $k\in[n]$ and any $t > \tau_k$, $x^t_{p_k} =
x^{\tau_k}_{p_k}$. This also means that none of the partial configurations
$x^t_{p^t}$ will change for $t > \tau_n$. Since none of the partial
configuration changed, $x^{\hat{\tau}_1} = x^{\tau_n}$. Since
$\CREG{p_1}{x^{\hat{\tau}_1}} = 0$ by assumption, then $x^{\hat{\tau}_1}$ is
conditionally optimal with respect to $p_1$. Likewise, $x^{\hat{\tau}_k} =
x^{\tau_n}$ and $\CREG{p_k}{x^{\hat{\tau}_k}} = 0$ for all $k\in[n]$, and thus
$x^T = x^{\hat{\tau}_n} = x^{\tau_n}$ is a local optimum.

\end{document}